\documentclass[10pt,twocolumn,letterpaper]{article}

\usepackage{cvpr}
\usepackage{times}
\usepackage{epsfig}
\usepackage{graphicx}
\usepackage{amsmath}
\usepackage{amssymb}

\usepackage[breaklinks=true,bookmarks=false]{hyperref}

\cvprfinalcopy 

\def\cvprPaperID{****} 

\begin{document}

\title{The Application of Two-level Attention Models in Deep Convolutional Neural Network for Fine-grained Image Classification}

\author{Tianjun Xiao$^1$  Yichong Xu$^2$ Kuiyuan Yang$^3$ Jiaxing Zhang$^3$ Yuxin Peng$^1$ Zheng Zhang$^4$\\
$^1$Institute of Computer Science and Technology, Peking University\\
$^2$Tsinghua University\\
$^3$Microsoft Research, Beijing\\
$^4$New York University Shanghai\\
{\tt\small xiaotianjun@pku.edu.cn, xycking@163.com, kuyang@microsoft.com}\\
{\tt\small jiaxz@microsoft.com, pengyuxin@pku.edu.cn, zz@nyu.edu}
}

\maketitle

\begin{abstract}
Fine-grained classification is challenging because categories can only be discriminated by subtle and local differences. Variances in the pose, scale or rotation usually make the problem more difficult. Most fine-grained classification systems follow the pipeline of finding foreground object or object parts (\emph{where}) to extract discriminative features (\emph{what}).

In this paper, we propose to apply visual attention to fine-grained classification task using deep neural network. Our pipeline integrates three types of attention: the bottom-up attention that propose candidate patches, the object-level top-down attention that selects relevant patches to a certain object, and the part-level top-down attention that localizes discriminative parts. We combine these attentions to train domain-specific deep nets, then use it to improve both the \emph{what} and \emph{where} aspects. Importantly, we avoid using expensive annotations like bounding box or part information from end-to-end. The weak supervision constraint makes our work easier to generalize.

We have verified the effectiveness of the method on the subsets of ILSVRC2012 dataset and CUB200\_2011 dataset. Our pipeline delivered significant improvements and achieved the best accuracy under the weakest supervision condition. The performance is competitive against other methods that rely on additional annotations.
\end{abstract}

\section{Introduction} \label{sec:introduction}

\begin{figure}[!htp]
\begin{center}
\includegraphics[width=0.48\textwidth]{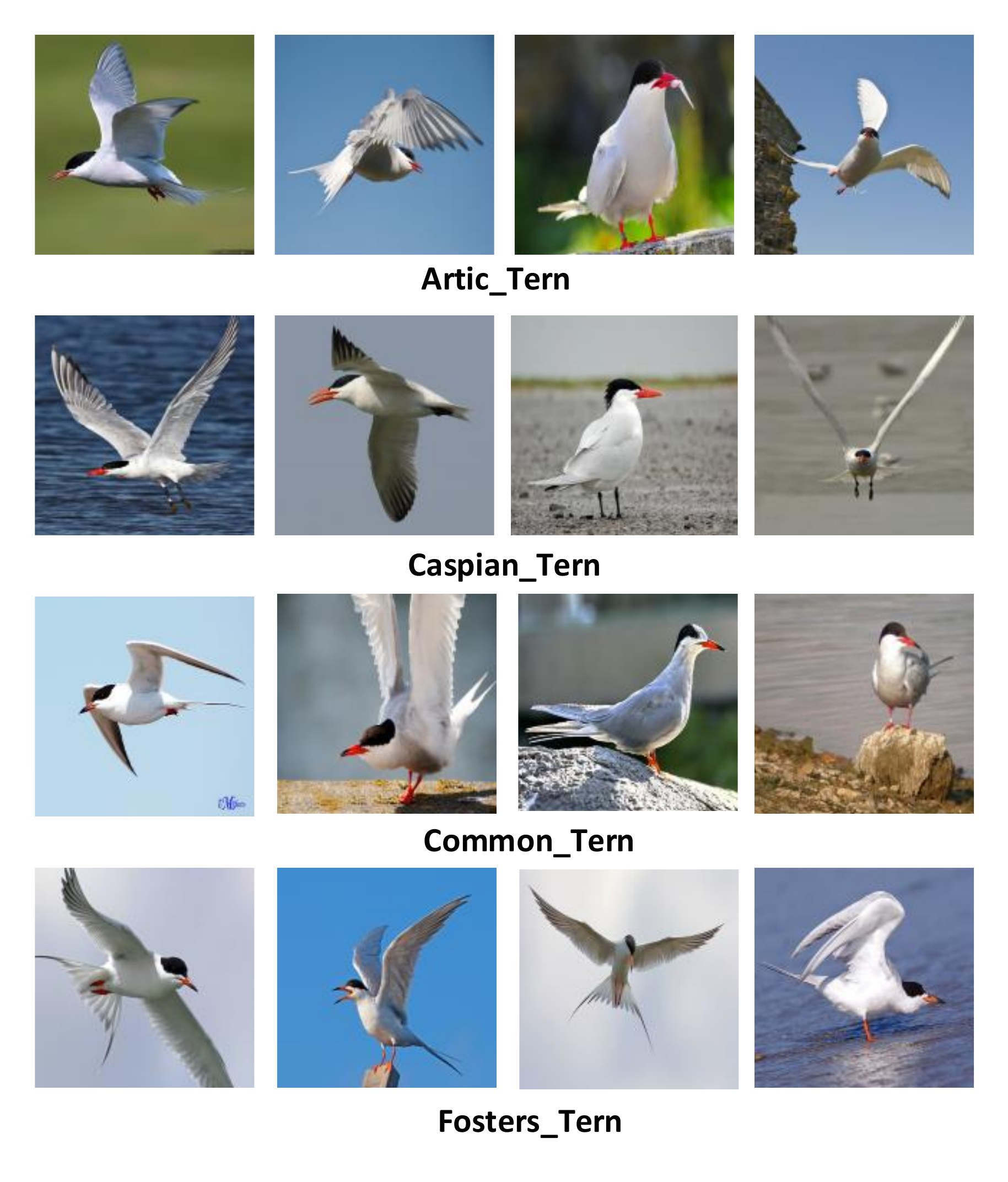}
\caption{Illustration of the difficulty of fine-grained classification : large intra-class variance and small inter-class variance.}
\label{fig:problem}
\end{center}
\end{figure}

Fine-grained classification is to recognize subordinate-level categories under some basic-level category, e.g., classifying different bird types~\cite{CUB200}, dog breeds~\cite{StanfordDog}, flower species~\cite{Flower}, aircraft models~\cite{aircraft} etc. This is an important problem with wide applications. Even in the ILSVRC2012 1K categories, there are 118 and 59 categories under the dog and bird class, respectively. Counter intuitively, intra-class variance can be larger than inter-class, as shown in Figure~\ref{fig:problem}. Consequently, fine-grained classification are technically challenging.

Specifically, the difficulty of fine-grained classification comes from the fact that discriminative features are  localized not just on foreground object, but more importantly on object parts~\cite{chai2013symbiotic} (e.g. the head of a bird). Therefore, most fine-grained classification systems follow the pipeline: finding foreground object or object parts (\emph{where}) to extract discriminative features (\emph{what}).

For this to work, a bottom-up process is necessary to propose image regions (or patches) that have high \emph{objectness}, meaning they contain parts of certain objects. Selective search ~\cite{uijlings2013selective} is an unsupervised process that can propose such regions at the order of thousands. This starting point is used extensively in recent studies~\cite{girshick14CVPR,zhang2014part}, which we adopt as well.

The bottom-up process has high recall but very low precision. If the object is relatively small, most patches are background and do not help classifying the object at all. This poses problems to the \emph{where} part of the pipeline, leading to the need of top-down attention models to filter out noisy patches and select the relevant ones. In the context of fine-grained classification, finding foreground object and object parts can be regarded as a two-level attention processes, one at object-level and another at part-level.

Most existing methods rely on strong supervision to deal with the attention problem. They heavily rely on human labels, using bounding box for object-level and part landmarks for part-level. The strongest supervision settings leverage both in training as well as testing phase, whereas the weakest setting uses neither. Most works are in between (see Section~\ref{sec:related} for an in-depth treatment).

Since labeling is expensive and non-scalable, the focus of this study is to use the weakest possible supervision. Recognizing the granularity differences, we employ two separate pipelines to implement object-level and part-level attention, but pragmatically leverage shared components. Here is a high level summary of our approach:

\begin{itemize}
	\item
	We turn a Convolutional Neural Net (CNN) pre-trained on ILSVRC2012 1K category into a \emph{FilterNet}. \emph{FilterNet} selects patches relevant to the basic-level category, thus processes the object-level attention. The selected patches drive the training of another CNN into a domain classifier, called \emph{DomainNet}.
	\item
	Empirically, we observe clustering pattern in the internal hidden representations inside the \emph{DomainNet}. Groups of neurons exhibit high sensitivity to discriminating parts. Thus, we choose the corresponding filters as \emph{part-detector} to implement part-level attention.
\end{itemize}
In both steps, we require only image-level labeling.

The next key step is to extract discriminative features from the regions/patches selected by these two attentions. Recently, there have been convincing evidence that features derived by CNN can deliver superior performance over hand-crafted ones ~\cite{zeiler2013visualizing,CNNfeat,DeCAF, zhang2014part}. Following the two attention pipelines outlined above, we adopt the same general strategies. At the object-level, the DomainNet directly output multi-view predictions driven by multiple relevant patches of an image. At the part-level, activations in the CNN hidden layers driven by detected parts yield another prediction through a \emph{part-based classifier}. The final classification merges results from both pipelines to utilize the advantage of the two level attentions.

Our preliminary results demonstrate the effectiveness of this design. With the weakest supervision, we improve the fine-grained classification in the dog and bird class of the ILSVRC2012 dataset from 40.1\% and 21.1\% to 28.1\% and 11.0\%, respectively. On the CUB200-2011~\cite{wah2011caltech} dataset, we reach 69.7\%, competitive to other methods that use stronger supervisions. Our technique improves naturally with better networks, for example the accuracy reaches nearly to 78\% using VGGNet~\cite{simonyan2014very}.

The rest of the paper is organized as follows. We first describe the pipeline utilizing object-level and part-level attentions for fine-grained classification in Section \ref{sec:methods}. Detailed performance study and analysis are covered in Section \ref{sec:experiment}. Related works are covered in Section \ref{sec:related}. Finally, We discuss what we learned, future work and conclusion in Section \ref{sec:conclusion}.

\begin{figure*}[t]
\begin{center}
\includegraphics[width=0.9\textwidth]{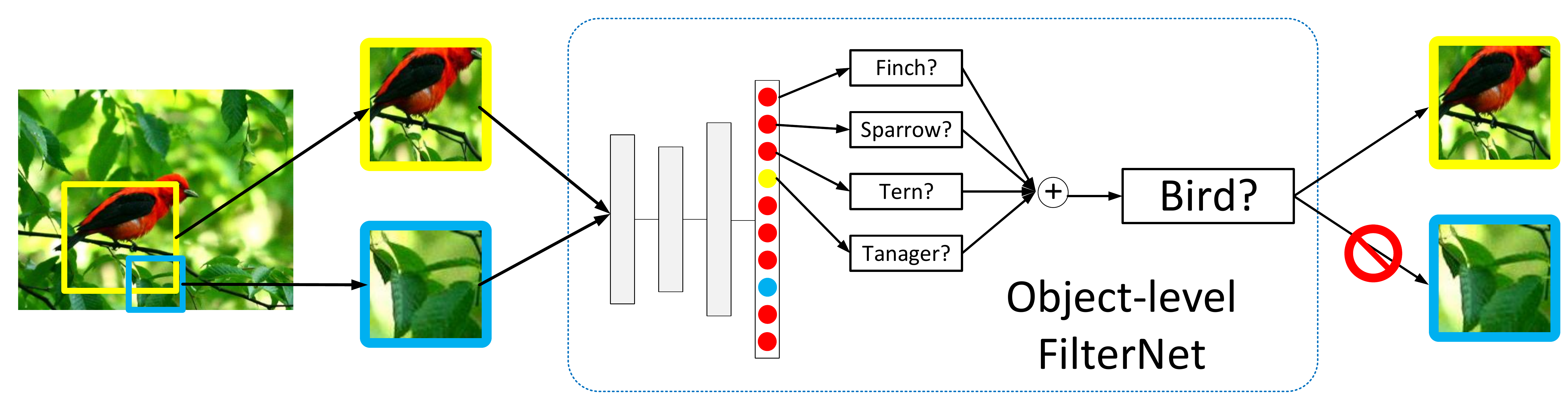}
\caption{Object-level top-down attention. An object-level FilterNet is introduced to decide whether to proceed a patch proposed by the bottom-up method to the next steps. The FilterNet only cares whether a patch is related to the basic level category, and targets filtering out background patches.}
\label{fig:objectatt}
\end{center}
\end{figure*}

\section{Methods} \label{sec:methods}

Our design is based on a very simple intuition: performing fine-grained classification requires first to ``see'' the object and then the most discriminative parts of it. Finding a Chihuahua in an image entails the process of first seeing a dog, and then focusing on its important features that tell it apart from other breeds of dog.

For this to work our classifier should \emph{not} work on the raw image but rather its constitute patches. Such patches should also retain the most objectness that are relevant to the recognition steps. In the example above, the objectness of the first step is at the level of dog class, and that of the second step is at the parts that would differentiate Chihuahua from other breeds (e.g. ear, head, tail). Crucially, recognizing the fact that detailed labeling are expensive to get and difficult to scale, we opt to use the weakest possible labels. Our pipeline uses only the image-level labels.

The raw candidate patches are generated in a bottom-up process, grouping pixels into regions that highlight the likelihood of parts of some objects. In this process, we adopt the same approaches as ~\cite{zhang2014part} and uses selective search ~\cite{uijlings2013selective} to extract patches (or regions) from input images. This step will provide multi-scale and multi-view of the original image. However, the bottom-up method will provide patches of high recall and low precision. Top-down attention need to be applied to select the relative patches useful for classification.

\subsection{Object-Level Attention Model}

\paragraph{Patch selection using object-level attention}
This step filters the bottom-up raw patches via a top-down, object-level attention. The goal is to remove noisy patches that are not relevant to the object. We do this by converting a CNN trained on the 1K-class ILSVR2012 dataset into a object-level \emph{FilterNet}. We summarize the activations of all the softmax neurons belonging to the parent class of a fine-grained category (e.g. for Chihuahua the parent class is the dog) as the selection confidence score, and then set a threshold on the score to decide whether a given patch should be selected. This is shown in Figure~\ref{fig:objectatt}. Through this way, the advantage of multi-scale and multi-view has been retained and also the noise has been filtered out.

\paragraph{Training a \emph{DomainNet}} 
The patches selected by the FilterNet are used to train a new CNN from scratch after proper warping. We call this second CNN the \emph{DomainNet} because it extracts features relevant to the categories belonging to a specific domain (e.g., dog, cat, bird).

We note that from a single image many such patches are made available, and the net effect is a boost of data augmentation. Unlike other data augmentation such as random cropping, we have a higher confidence that the patches are relevant. The amount of data also drives training of a bigger network, allowing it to build more features. This has two benefits. First, the DomainNet is a good fine-grained classifier itself. Second, its internal features now allow us to build part detectors, as we will explain next.

\paragraph{Classification using object-level attention} 
The patch selection using object-level attention can be naturally applied to the testing phase. To get the predicted label of an image, we provide the DomainNet with the patches selected by the FilterNet to feed forward. Then compute the average classification distribution of the softmax output for all the patches. Finally we can get the prediction on the averaged softmax distribution.

\begin{figure*}[!htp]
	\begin{center}
		\includegraphics[width=0.9\textwidth]{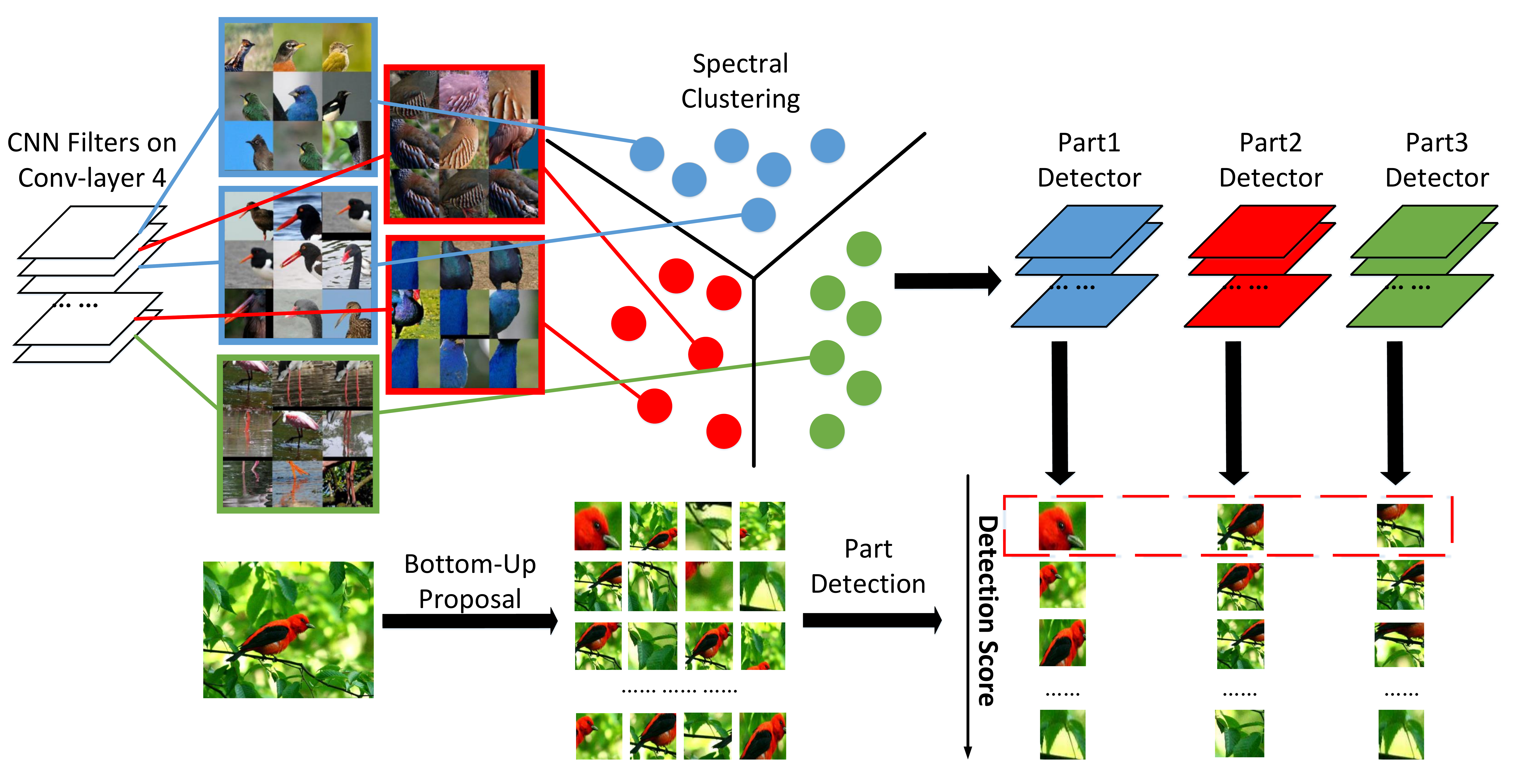}
		\caption{Part-level top-down Attention: The filters in the DomainNet shows special interests on specific object parts and clustering pattern can be found among filters according to their interested parts. We use spectral clustering to find the groups, then use the filters in a group to serve as part detector. In this figure, mid-level CNN filters can be served as head detector, body detector and leg detector for birds.}
		\label{fig:partatt}
	\end{center}
\end{figure*}

\subsection{Part-Level Attention Model}

\paragraph{Building the part detector} The work of DPD~\cite{zhang2013deformable} and Part-RCNN~\cite{zhang2014part} strongly suggest that certain discriminative local features (e.g. head and body) are critical to fine-grained classification. Instead of using the strong labels on parts and key points, as is done in many related works~\cite{zhang2013deformable, zhang2014part, branson2014bird}, we are inspired by the fact that hidden layers of the DomainNet have shown clustering patterns. For example, there are groups of neurons respond to bird head, and others to bird body, despite the fact they may correspond to different poses. In hindsight, this is not at all surprising, given that these features indeed ``stand out'' and ``speak for'' a category.

Figure~\ref{fig:partatt} shows conceptually what this step performs. Essentially, we perform spectral clustering on the similarity matrix $S$ to partition the filters in a middle layer into $k$ groups, where $S(i, j)$ denotes the cosine similarity of the weights of two mid-layer filters $F_i$ and $F_j$ in the DomainNet. In our experiments, our network is essentially the same as the AlexNet~\cite{AlexNet}, and we pick neurons from the 4th convolution layer with $k$ set to 3. Each cluster acts as a part detector. Raw patches extracted from the image (using selective search) are warped to fit the receptive size of these neurons, and the clusters now vote and generate a final detection score.

\begin{figure}[!htp]
	\begin{center}
		\includegraphics[width=0.45\textwidth]{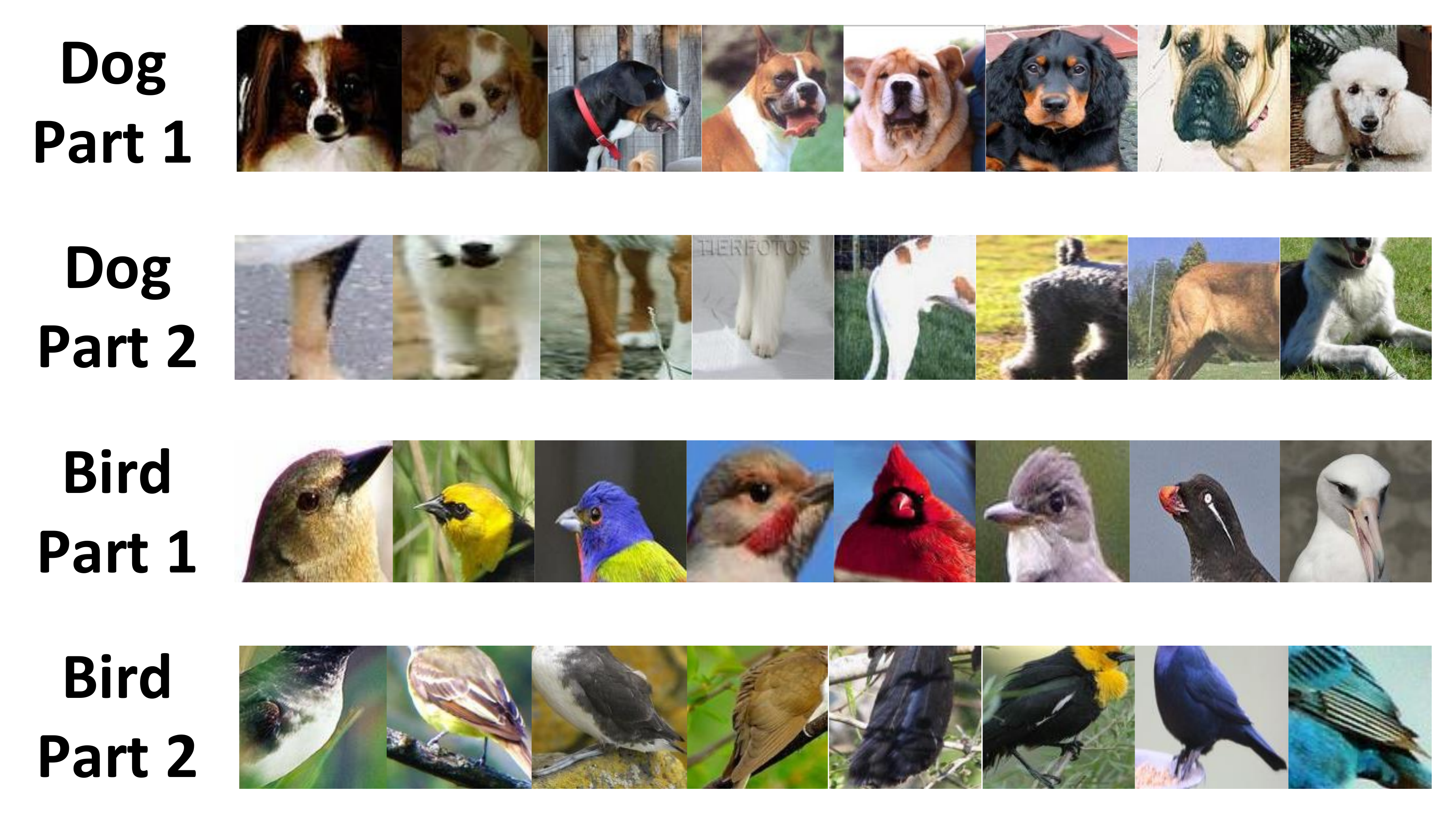}
		\caption{Part-level top-down attention detection results. One group of filters in bird DomainNet pay specially attention to bird head, and the other group to bird body. Similarly, for the dog DomainNet, one group of filters pay attention to dog head, and one to dog legs}
		\label{fig:detectres}
	\end{center}
\end{figure}

Some detection results of the dog and bird class are shown on Figure~\ref{fig:detectres}. It's clear that one group of filters in bird DomainNet pay specially attention to bird head, and the other group to bird body. Similarly, for the dog DomainNet, one group of filters pay attention to dog head, and one to dog legs. According to our practicing experience, filters detect background noise will be clustered into one group. We can use the filter visualization technique or use a validation set at testing time to find it out to prevent the noise disturbing performance.

\paragraph{Building the part-based classifier} 
The patches that selected by the part detector generate activations inside the DomainNet. We concatenate these activations and train a SVM as the part-based classifier.

\begin{figure*}[!htp]
	\begin{center}
		\includegraphics[width=1.0\textwidth]{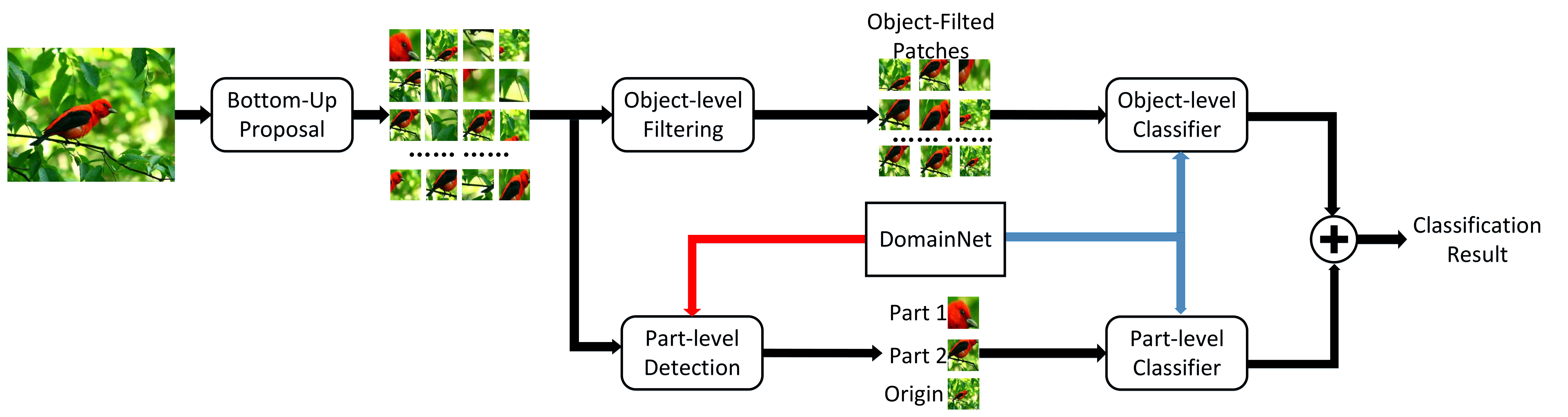}
		\caption{The complete classification pipeline of our method. Two levels of top-down attentions are applied on the bottom-up proposals. One conducts object-level filtering to select patches relevant to bird to feed into the classifier. The other conducts part-level detection to detect parts for classification. DomainNet can provide the part detectors for part-level method and also the feature extractor for both of the two level classifiers.  The prediction results of the two classifiers are merged in later phase to combine the advantages of the two level attentions.}
		\label{fig:whole_pipeline}
	\end{center}
\end{figure*}

\subsection{The Complete Pipeline}
The domainNet classifier and the part-based classifier are both fine-grained classifiers. However, their functionality and strength differ, primarily because they admit patches of different nature. The bottom-up process using selective search are raw patches. From them, the FilterNet selects multiple views that (hopefully) focus on the object as a whole; these patches drive the DomainNet. On the other hand, the part-based classifier selects and works exclusively on patches containing discriminate and local features. Finally, we merge the prediction results of the two level attention methods to utilize the advantage of the two. Even though some patches are admitted by both classifiers, their features have different representation in each and can potentially enrich each other.

Figure~\ref{fig:whole_pipeline} shows the complete pipeline and when we merge the results of the two level attention classifiers.

\section{Experiment} \label{sec:experiment}

This section presents performance evaluations and analysis of our proposed method on three fine-grained classification tasks:
\begin{itemize}
	\item
	Classification of two subsets in ILSVRC2012, the dog dataset (ILSVRC2012\_Dog) and the bird dataset (ILSVRC2012\_bird). The firs contains 153,773 images of 118 breeds of dog, and the second contains 79,491 images of 59 types of bird. The train/test split follows standard protocol of ILSVRC2012. Both datasets are weakly annotated, where only class labels are available. 
	\item
	The widely-used fine-grained classification benchmark Caltech-UCSD Birds dataset~\cite{wah2011caltech} (CUB200-2011), with 11,788 images of 200 types of birds. Each Image in CUB200-2011 has detailed annotations, including image level label, bounding box and part landmarks. 
\end{itemize}

\subsection{Implementation Details}
Our CNN architecture is the popular Alex Net \emph{et al.}~\cite{AlexNet}, with 5 convolutional layers and 3 fully connected layers. It is used in all experiments, except the number of neurons of the output layer is set as number of categories when required. For a fair comparison, we try to reproduce results of other approaches on the same network architecture. When using CNN as feature extractor, the activations of the first fully-connected layer are outputted as features. Finally, to demonstrate that our method is agnostic to network architecture and can improve with it, we also try to use the more recent VGGNet~\cite{simonyan2014very} in the feature extraction phase. Due to time limit, we have not replicated all results using the VGGNet.

\subsection{Results on ILSVRC2012\_Dog/Bird}
In this task, only image-level class labels are available. Therefore, fine-grained methods requiring detailed annotations are not applicable. For brevity, we will only report results on dog; results of bird are qualitatively similar.

The baselines are performance of CNN but trained with two different strategies, including:
\begin{itemize}
    \item CNN\_domain: The network is trained only on images from dog categories. In the training phase, randomly cropped $224\times224$ patches from the whole image are used to avoid overfitting. In testing phase, softmax outputs of 10 fixed views (the center patch, the four corner patches, and their horizonal reflections) are averaged as the final prediction. In this method, no specific attention is used and patches are equally selected.
    \item CNN\_1K: The network is trained on all images of ILSVRC2012 1K categories, then the softmax neurons not belong to dog are removed. Other settings are the same as above. This is a multi-task learning method that simultaneously learns all models, including dog and bird. This strategy utilizes more data to train a single CNN, and resist overfitting better, but has the tradeoff of wasting capacity on unwanted categories.
\end{itemize}
These baseline numbers are compared with three strategies of our approach: using object-level and part-level attention only, and the combination of both. Selective search proposes several hundred number of patches, and we let FilterNet to select roughly 40 of them, using a confidence score of 0.9.

Table~\ref{tab:ImageNet} summarizes the top-1 error rates of all five strategies. It turns out the two baselines perform about the same. However, our attention based methods achieves much lower error rates. Using object-level attention only drops the error rate by 9.3\%, comparing against CNN trained with randomly cropped patches. This clearly demonstrates the effectiveness of object-level attention: the DomainNet now focuses on learning domain specific features from foreground objects. Combining part-level attention, the error rate drops to 28.1\%, which is significantly better than the baselines. The result of using part-level attention alone is not as good as object-level attention, as there are still more ambiguities in part level. However, it achieves pose normalization to resist large pose variations, and is complementary to the object-level attention.

\begin{table*}[htp]
\caption{Top-1 error rate on ILSVRC2012\_Dog/Bird validation set.}
\centering 
\begin{tabular}{l|c|c} 
\hline
Method & ILSVRC2012\_Dog & ILSVRC2012\_Bird \\ \hline
CNN\_domain & 40.1 & 21.1 \\
CNN\_1K & 39.5 & 19.2 \\
Object-level attention & 30.3 & 11.9 \\
Part-level attenion & 35.2 & 14.6 \\
Two-level attention & \textbf{28.1} & \textbf{11.0}  \\ \hline

\end{tabular}
\label{tab:ImageNet}
\end{table*}

\subsection{Results on CUB200-2011}
For this task, we begin with a demonstration of the performance advantage of learning deep feature based on object level attention. We then present full results against other state-of-the-art methods.

\begin{figure}[htp]
\begin{center}
\includegraphics[width=0.45\textwidth]{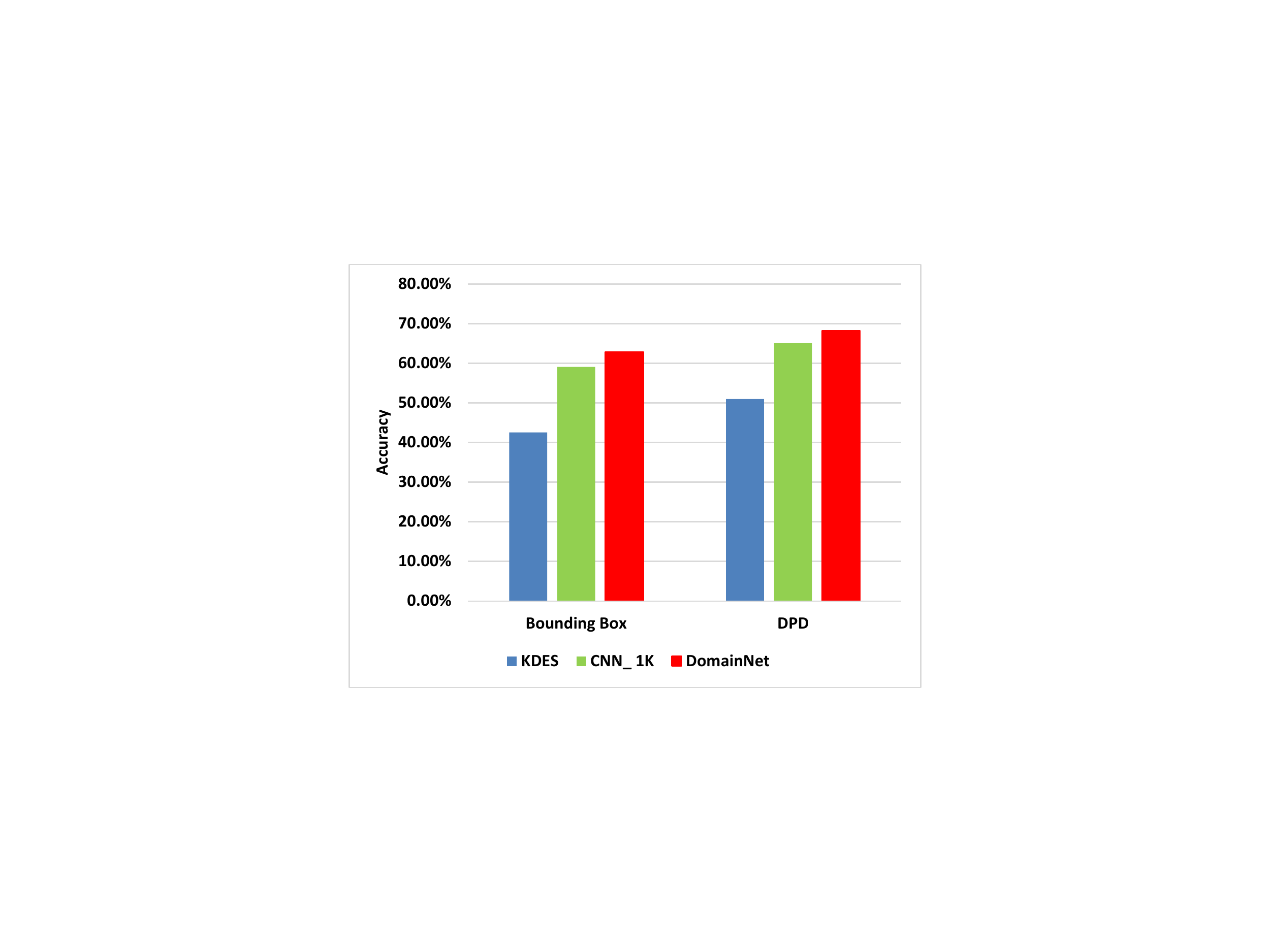}
\caption{Comparision of different feature extractors by attentions provided by bounding box and DPD.}
\label{fig:featuredif}
\end{center}
\end{figure}

\paragraph{Advantage on Learning Deep Feature}
We have shown that the bird DomainNet trained with object-level attention delivers superior classification performance on ILSVRC2012\_Bird. It is reasonable to assume that part of the gain comes from the better learned features. In this experiment, we use the domainNet as feature extractor on CUB200-2011 to verify the advantage of those features. We compare against two baseline feature extractors, one is hand-crafted kernel descriptors~\cite{bo2010kernel} (KDES) which was widely used in fine-grained classification before using CNN feature, the other is the CNN feature extractor pre-trained from all the data in ILSVRC2012~\cite{CNNfeat}. We compared the feature extractors under two classification pipelines and the features are fed in a SVM classifier. The first one uses bounding boxes, the second one is proposed in Zhang \emph{el al.}~\cite{zhang2013deformable} (DPD) which relies on deformable part based detector~\cite{felzenszwalb2010object} to find object and its parts. In this experiment, no CNN is finetuned on CUB200-2011. As shown in Figure~\ref{fig:featuredif}, DomainNet based feature extractor achieves the best results on both pipelines. This further demonstrates that using object-level attention to filter relevant patches is an important condition for CNN to learn good features.

\paragraph{Advantage of the Classification Pipeline}
In this experiment, the DomainNet is fine-tuned using CUB200-2011 with patches generated by object-level attention. The mean accuracies are reported in Table~\ref{tab:OPTestCUB}, along with how much annotations are used. These methods are grouped into three sets. The first set is our attention-based methods, the second uses the same DomainNet feature extractor as the first set but with different pipeline and annotations, and the third set includes the state-of-the-art results from recent literatures. 

We first compare the results of the first two set where the used feature extractor is the same, and the performance difference is attributed to different attention models. Using original image only achieves the lowest accuracy (58.8\%).
which demonstrates the importance of object and part level attention in fine-graind image classification. 
In comparison, our attention-based methods achieved significant improvement, and the two-level attention delivers even better results than using human labelled bounding box (69.7\% vs. 68.4\%), and is comparable to DPD (70.5\%). The DPD result is based on implementation using our feature extractor, it used deformable part-based detector trained with object bounding box. The standard DPD pipeline also need bounding box at testing time to produce relatively good performance. To the best of our knowledge, 69.7\% is the best result under the weakest supervision.

The third set summarizes the state-of-the-art methods.  Our results is much better than the ones using only bounding boxes in training and testing, but still has gap to the methods using part-level annotation. 

\begin{table*}[htp]
\caption{Accuracy Comparison between methods using bounding box} 
\centering 
\begin{tabular}{l|c|c|c|c|c} 
\hline
 & \multicolumn{2}{c|}{Training phase} & \multicolumn{2}{c|}{Testing phase} &  \\ \hline
Method & BBox Info &  Part Info & BBox Info & Part Info & Accuracy (\%) \\ \hline
Object-level attention & & & & & 67.6  \\ \hline
Part-level attention  & & & & & 64.9  \\ \hline
Two-level attention & & & & & \textbf{69.7}  \\ \hline \hline

DomainNet without attention & & & & & 58.8 \\ \hline
BBox + DomainNet & & & \checkmark & & 68.4  \\ \hline
DPD~\cite{zhang2013deformable} + DomainNet & \checkmark & & \checkmark & & 70.5  \\ \hline \hline

Symbiotic~\cite{chai2013symbiotic} & \checkmark &  & \checkmark & & 61.0  \\ \hline
Alignment~\cite{gavves2013fine} & \checkmark &  & \checkmark & & 62.7  \\ \hline
DeCAF$_6$~\cite{DeCAF} & \checkmark & & \checkmark & & 58.8 \\ \hline
CNNaug-SVM~\cite{CNNfeat} & \checkmark & & \checkmark & & 61.8 \\ \hline
Part RCNN~\cite{zhang2014part} & \checkmark & \checkmark & & & 73.5  \\ \hline
Pose Normalized CNN~\cite{branson2014bird} & \checkmark & \checkmark &  & & 75.7  \\ \hline
POOF~\cite{berg2013poof} & \checkmark & \checkmark & \checkmark &  & 56.8  \\ \hline
Part RCNN~\cite{zhang2014part} & \checkmark & \checkmark & \checkmark & & 76.7 \\ \hline
POOF~\cite{berg2013poof} & \checkmark & \checkmark & \checkmark & \checkmark & 73.3 \\ \hline
\end{tabular}
\label{tab:OPTestCUB}
\end{table*}

Our results can be improved by using more powerful feature extractors. If we use the VGGNet~\cite{simonyan2014very} to extract feature, the baseline method without attention by only using original image can be improved to 72.1\%. Adding object-level attention, part-level attention, and the combined attentions boost the performance to 76.9\%, 76.4\% and 77.9\%, respectively.

\section{Related Work} \label{sec:related}
Fine-grained classification has been extensively studied recently~\cite{wah2011caltech, CUB200, StanfordDog, bo2010kernel, chai2013symbiotic, yao2012codebook, zhang2013deformable, berg2013poof, branson2014bird}. Previous works have aimed at boosting the recognition accuracy from three main aspects: 1. object and part localization, which can also be treated as object/part level attention; 2. feature representation for detected objects or parts; 3. human in the loop~\cite{wah2011multiclass}. Since our goal is automatic fine-grained classification, we focus on the related work of the first two.

\subsection{Object/Part Level Attention}
In fine-grained classification tasks, discriminative features are mainly localized on foreground object and even on object parts, which makes object and part level attention be the first important step. As fine-grained classification datasets are often using detailed annotations of bounding box and part landmarks, most methods rely on some of these annotations to achieve object or part level attention. 

The strongest supervised setting is using bounding box and part landmarks in both training and testing phase, which is often used to test performance upbound~\cite{berg2013poof}. To verify CNN features on fine-grained task, bounding boxes are assumed given in both training and testing phase~\cite{DeCAF,CNNfeat}. Using provided bounding box, several methods proposed to learn part detectors in unsupervised or latent manner~\cite{yang2012unsupervised,chai2013symbiotic}. To further improve the performance, part level annotation is also used in training phase to learn strongly-supervised deformable part-based model~\cite{azizpour2012strongdpm,zhang2013deformable} or directly used to finetune pre-trained CNN~\cite{branson2014bird}.

Our work is also closely related to recently proposed object detection method (R-CNN) based on CNN feature~\cite{girshick14CVPR}. R-CNN works by first proposing thousands candidate bounding boxes for each image via some bottom-up attention model~\cite{uijlings2013selective,cheng2014bing}, then selecting the bounding boxes with high classification scores as detection results.
Based on R-CNN, Zhang \emph{et al.} has proposed Part-based R-CNN~\cite{zhang2014part} to utilize deep convolutional network for part detection. 

\subsection{Feature Representation}
The other aspect to directly boost up the accuracy is to introduce more discriminative feature to represent image regions. Ren \emph{et al.} has proposed Kernel Descriptors~\cite{bo2010kernel} and were widely used in fine-grained classification pipelines~\cite{zhang2013deformable,yang2012unsupervised}. Some recent works try to learn feature descriptions from the data, Berg \emph{et al.} has proposed the part-based one-vs-all features library POOF~\cite{berg2013poof} as the mid-level features. CNN feature extractors pre-trained from ImageNet data also showed significant performance improvement on fine-grained datasets~\cite{CNNfeat, DeCAF}. Zhang \emph{et al.} further improved the performance of CNN feature extractor by fine-tuning on fine-grained dataset~\cite{zhang2014part}. 

Our approach adopts the same general principle. We also share the same strategy of taking region proposals in a bottom-up process to drive the classification pipeline, as is done in R-CNN and Part R-CNN. One difference is that we enrich the object-level pipeline with relevant patches that offer multiple views and scales. More importantly, we opt for the weakest supervision throughout the model, relying solely on CNN features to implement attention, detect parts and extract features. 

\section{Conclusions} \label{sec:conclusion}

In this paper, we propose a fine-grained classification pipeline combining bottom-up and two top-down attentions. The object-level attention feeds the network with patches relevant to the task domain with different views and scales. This leads to better CNN feature for fine-grained classification, as the network are driven by domain-relevant patches that are also rich with shift/scale variances. The part-level attention focuses on local discriminate patterns and also achieves pose normalization. Both levels of attention can bring significant gains, and they compensate each other nicely with late fusion. One important advantage of our method is that, the attention is derived from the CNN trained with classification task, thus it can be conducted under the weakest supervision setting where only class label is provided. This is in sharp contrast with other state-of-the-art methods that require object bounding box or part landmark to train or test. To the best of our knowledge, we get the best accuracy on CUB200-2011 dataset under the weakest supervision setting.

These results are promising. At the same time, the experience points out a few lessons and future directions, which we summarize as the followings:
\begin{itemize}
\item Dealing with ambiguities in part level attention. Our current method does not fully utilize what has been learned in CNN. Filters of different layers should be considered as a whole to facilitate robust part detection, since part feature may appear in different layers due to the scale issue.

\item A closer integration of the object-level and part-level attention. One advantage of object-level attention is that it can provide large amount of relevant patches to help resist variance to some extend. However, this is not leveraged by the current part-level attention pipeline. We may borrow the idea of multi-patch testing to part-level attention method to derive more effective pose normalization.
\end{itemize}

We are actively pursuing the above directions.

\end{document}